\documentclass[runningheads]{llncs}

\usepackage{eccv}

\usepackage{algorithm}
\usepackage{algorithmic}

\usepackage{eccvabbrv}

\usepackage{graphicx}
\usepackage{booktabs}

\usepackage[accsupp]{axessibility}  

\usepackage{hyperref}

\usepackage{orcidlink}

\begin{document}
	
	\title{DeltaDeno: Zero-Shot Anomaly Generation via Delta-Denoising Attribution} 
	
	\titlerunning{DeltaDeno}
	
	\author{
		Chaoran Xu\inst{1,2}\orcidlink{0009-0007-2025-3784}
		\and
		Chengkan Lv\inst{1,2,3}\orcidlink{0000-0001-5319-1363}
		\and
		Qiyu Chen\inst{1,2}\orcidlink{0009-0003-7910-862X}
		\and
		Yunkang Cao\inst{4}\orcidlink{0000-0001-7619-6618}
		\and
		Feng Zhang\inst{1,2,3}\orcidlink{0000-0002-8399-6143}\thanks{Corresponding author.}
		\and
		Zhengtao Zhang\inst{1,2,3}\orcidlink{0000-0003-1659-7879}
	}
	
	\authorrunning{C. Xu et al.}
	
	\institute{
		Institute of Automation, Chinese Academy of Sciences, Beijing, China
		\and
		School of Artificial Intelligence, University of Chinese Academy of Sciences, Beijing, China
		\and
		CASI Vision Technology CO., LTD., Luoyang, China
		\and
		School of Artificial Intelligence and Robotics, Hunan University, Changsha, China
		\email{\{xuchaoran2024,chengkan.lv,chenqiyu2021,feng.zhang,zhengtao.zhang\}@ia.ac.cn}
		\email{caoyunkang@ieee.org}
	}
	
	\maketitle

	\begin{abstract}
		Anomaly generation is often framed as few-shot fine-tuning with anomalous samples, which contradicts the scarcity that motivates generation and tends to overfit category priors. We tackle the setting where no real anomaly samples or training are available. We propose Delta-Denoising (\textbf{DeltaDeno}), a training-free zero-shot anomaly generation method that localizes and edits defects by contrasting two diffusion branches driven by a minimal prompt pair under a shared schedule. By accumulating per-step denoising deltas into an image-specific localization map, we obtain a mask to guide the latent inpainting during later diffusion steps and preserve the surrounding context while generating realistic local defects. To improve stability and control, DeltaDeno performs token-level prompt refinement that aligns shared content and strengthens anomaly tokens, and applies a spatial attention bias restricted to anomaly tokens in the predicted region. Experiments on public datasets show that DeltaDeno achieves great generation, realism and consistent gains in downstream detection performance. 
		Code will be made publicly available at \url{https://github.com/CROVO1026/DeltaDeno}.
		
		\keywords{ Anomaly generation \and Anomaly detection  \and Zero-shot}
	\end{abstract}

	\section{Introduction}
	\label{sec:intro}
	
	Automated visual inspection~\cite{bergmann2019mvtec} is a cornerstone of modern smart manufacturing and safety-critical operations. In principle, high-performance anomaly detection (AD) systems can be trained with sufficient defective data under supervised learning~\cite{bergmann2020uninformed,chen2025center,quzhen2023investigating,chenqiyu2024unified,xu2026mrad,anomalygpt}. In practice, defects are rare, diverse, and costly to curate. Early-stage production lines and high-yield semiconductor processes exhibit exceptionally low defect rates. Many defects emerge only after process changes or tool aging and pixel-accurate masks also require expert annotation~\cite{fang2025boosting,zhou2023anomalyclip,anomalyany,anomalydiffusion}. Beyond scarcity, domains evolve over time through changes in materials, suppliers, and lighting, which creates persistent distribution shifts that degrade deployed AD~\cite{dual}. These factors motivate \emph{anomaly generation (AG)} as a complementary route to provide realistic and controllable abnormal samples and to stress-test detectors without halting production~\cite{anomalydiffusion,defectfill,draem}.
	
	Early AG efforts centered on cut-paste composition~\cite{croppaste,draem,cutpaste} and GAN-based generators~\cite{dfmgan,defectgan}. With the rapid advances in diffusion models~\cite{ldm,ddpm,iddpm,classifierfree,diffusionbeat}, diffusion-based AG methods have since emerged and become a major line of work that are usually guided by masks, bounding boxes, or short text prompts~\cite{anomalyany,shi2024few,seas}. This line of research is primarily driven by two objectives. The first is \emph{realism}~\cite{defectfill,dual}, meaning that the generated region should be photometrically and geometrically consistent with the background. The second is \emph{controllability}~\cite{shi2024few}, meaning that one can specify where the change occurs, how strong it is, and what type of defect is inserted.
	\begin{figure}[t]
		\centering
		\includegraphics[width=\linewidth]{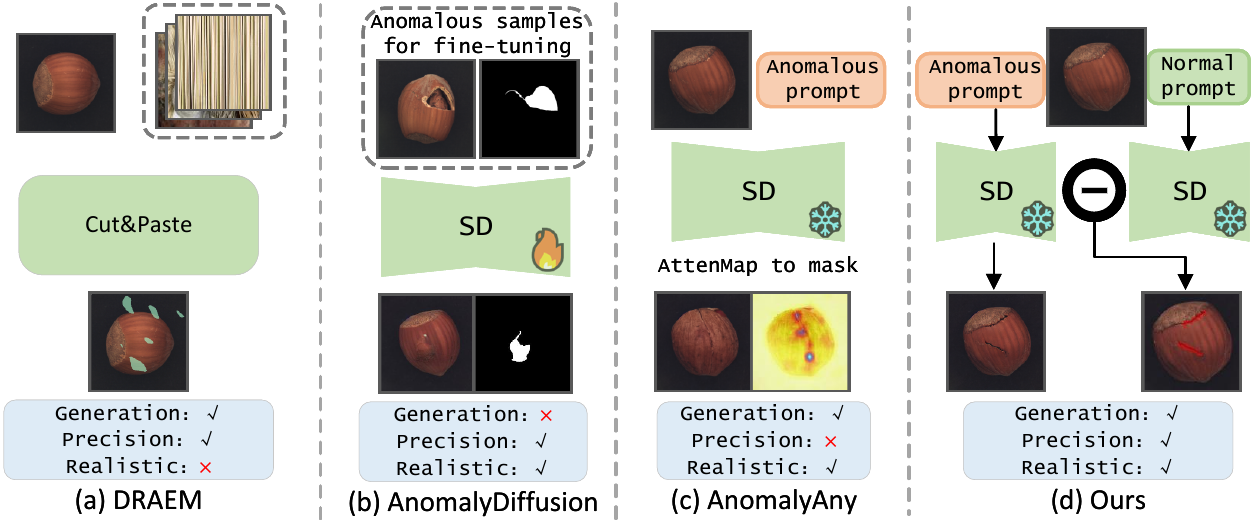}
		\caption{\textbf{Comparison between visual anomaly generation methods.}
			Compared with prior approaches, \textbf{DeltaDeno} delivers cross-category generalization, precise masks, and high realism while requiring no fine-tuning.}
		\label{fig:method_compare}
	\end{figure}
	
	Despite steady progress, many current pipelines adopt \emph{few-shot anomaly generation (FSAG)}~\cite{defectfill,dfmgan,anomalydiffusion,shi2024few,dual}. A pretrained generator is adapted with a small number of real anomalous samples and masks, and the adapted model is then used to expand the training set. This practice runs against the motivation for AG, which is to reduce dependence on rare and costly annotations. In many deployments, anomalous samples do not exist at scale, are expensive to collect, or are restricted by safety and privacy rules. FSAG also binds the generator to category-specific priors such as texture, lighting, and geometry, which encourages overfitting and weakens transfer across categories. As a result, gains on category A often fail to carry over to category B without anomalous samples.
	\emph{Zero-shot anomaly generation (ZSAG)} aims to remove this dependence on anomalous pixels and additional training. AnomalyAny~\cite{anomalyany} is a representative ZSAG example that shows good cross-category generalization, yet it also exposes practical limits. As shown in Fig.~\ref{fig:method_compare}, DRAEM follows the cut\mbox{-}paste paradigm and shows strong generalization ability but very limited realism~\cite{draem}. 
	The representative FSAG approach, AnomalyDiffusion, relies on supervised fine\mbox{-}tuning with real anomalous samples and masks, which makes it difficult to generalize when no downstream anomalies are available~\cite{anomalydiffusion}. 
	AnomalyAny builds masks from attention maps, but their low resolution and inaccurate localization remain problematic.

	To address these limitations, we present \emph{DeltaDeno}, a training-free ZSAG framework that localizes and generates defects without any anomalous samples or model adaptation. We use a minimal pair of prompts that differ only by an anomaly descriptor to drive two synchronized diffusion branches under the same schedule. At each step, we compare the two branches’ denoiser predictions and accumulate their differences to build an image-specific anomaly attribution map as mask. Later, we perform mask-guided latent inpainting, which refines the initially coarse anomaly regions during later time-steps while preventing the synthesized defects from spilling into normal areas.
	We also apply \textbf{normal-reference guided initialization}, which starts from a partially noised encoding of the normal image to keep synthesis aligned with the original content distribution and to shorten the reverse trajectory.
	To further sharpen localization, we introduce lightweight \textbf{token-level prompt refinement} and \textbf{spatial attention biasing}, which strengthen the anomaly-token discrepancy across the two denoising branches and suppress the influence of non-anomaly tokens, yielding a more precise mask with fewer spurious edits. This attribution-guided design ties localization to the model’s own denoising dynamics and produces sharper, more realistic anomalies without fine-tuning. The main contribution of this paper can be summarized as follows:
	\begin{itemize}
		\item We propose a zero-shot, training-free anomaly generation framework DeltaDeno that requires no anomalous samples, no fine-tuning, and no task-specific retraining.
		\item We introduce \emph{delta-denoising attribution} for ZSAG, using stepwise denoising discrepancies between normal and anomaly prompts to localize and generate defects, while lightweight prompt refinement and spatial attention biasing improve realism and mask precision.
		\item We demonstrate strong localization fidelity, photorealism, and cross-category generalization across industrial and other domains, yielding competitive improvements in downstream AD performance.
	\end{itemize}
	
	\section{Related Work}
	\label{sec:related_work}

	\subsection{Diffusion-based Image Editing}
	
	Text-guided diffusion enables high-quality and wide-coverage image editing, with Stable Diffusion~\cite{ldm} and its latent variants serving as scalable backbones for reconstruction and control. 
	Some methods focus on preserving structural consistency while steering appearance. 
	For example, SDEdit~\cite{sdedit} adjusts denoising schedules to balance fidelity and edit strength, while Prompt-to-Prompt~\cite{prompt2p} constrains cross-attention so that edits follow token semantics without disturbing the spatial layout. 
	Attend-and-Excite~\cite{attendandexcite} further amplifies weak tokens to improve semantic faithfulness. 
	Other approaches enhance localization by tightening the reconstruction path, as in Null-text inversion~\cite{nulltext} and Imagic~\cite{imagic}, or by confining edits with masks as demonstrated in blended latent diffusion~\cite{blended}. 
	DiffEdit~\cite{diffedit} derives an edit mask by contrasting source and target prompts under a shared schedule, while subsequent methods such as MasaCtrl~\cite{masactrl} and InstructPix2Pix~\cite{instructpix2pix} extend diffusion control to multi-prompt and instruction-driven editing. 
	Inspired by these works, we edit normal images with anomaly semantics, enabling training-free, context-consistent anomaly generation.

	\subsection{Anomaly Generation }
	Anomaly generation provides synthetic abnormal samples for training, evaluation, and stress testing of anomaly detection systems. Early practical approaches rely on cut-paste composition, such as DRAEM, which transfers defect patches onto normal images but often suffers from lighting and material inconsistency~\cite{draem}. GAN-based AG~\cite{dfmgan,defectgan} improves texture realism and diversity through adversarial learning, while diffusion-based backbones further enhance fidelity and controllability by introducing masks, boxes, or text prompts to guide synthesis. A common trend is FSAG~\cite{anomalydiffusion,seas,dual}, which fine-tunes a pretrained generator on a small set of real anomalies and pixel masks to expand training data. Although FSAG produces detailed and category-specific defects, it depends on scarce annotations and tends to overfit to class priors, limiting cross-domain generalization. To remove this dependence, ZSAG leverages the intrinsic guidance of pretrained diffusion models and text prompts. Recent ZSAG methods, such as AnomalyAny~\cite{anomalyany} and AnoStyler~\cite{anostyler}, broaden category coverage, yet still face limitations in generation realism and mask precision to varying degrees. In contrast,  we aggregate step-wise denoising discrepancies between normal and anomaly prompts to directly attribute semantic divergence, enabling training-free,  high-quality local anomaly generation with matched mask.

	\section{Method}
	\label{sec:method}
\begin{figure}[t]
		\centering
		\includegraphics[width=\linewidth]{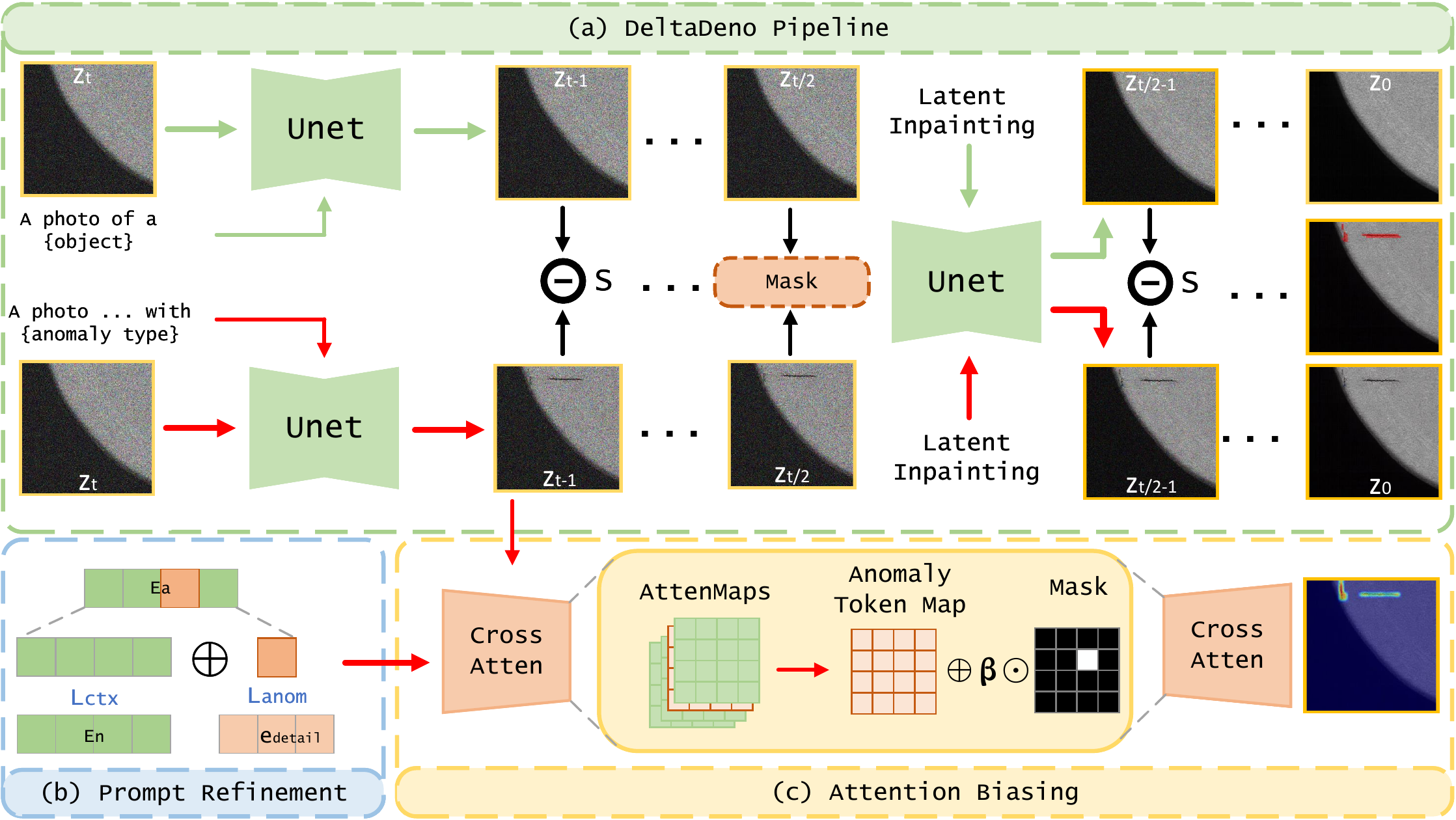}
		\caption{\textbf{Overview of the DeltaDeno framework integrating delta-denoising localization, prompt refinement, and attention biasing.} 
		}
		\label{fig:model_pipline}
	\end{figure}

	\noindent
	We present \emph{DeltaDeno}, a training-free zero-shot anomaly generation framework that localizes and generates defects through delta-denoising  attribution, without any anomalous pixel labels or model adaptation, as shown in  Fig.~\ref{fig:model_pipline}.
	We contrast the denoiser predictions of two synchronized diffusion branches, driven by normal and anomaly prompts, to capture semantic divergences as anomaly masks. 
	We then refine these coarse regions via mask-guided latent inpainting, preserving contextual consistency and preventing spill-over into clean areas. 
	To maintain alignment with the normal content distribution, we apply \textbf{normal reference guided initialization}, which starts from a partially noised normal encoding and shortens the reverse trajectory. 
	Furthermore, lightweight \textbf{token-level prompt refinement} and \textbf{spatial attention biasing} sharpen localization by emphasizing anomaly-related tokens and suppressing irrelevant ones.

	\subsection{Preliminaries}
	
	\noindent\textbf{Latent diffusion models.}
	Denoising diffusion models learn the data distribution by predicting the noise added at each step. 
	We follow latent diffusion, where an encoder $\mathcal{E}$ maps an image $x$ to a latent $z_0=\mathcal{E}(x)$ and a decoder $\mathcal{D}$ reconstructs images from latents~\cite{vae}. 
	Given a noise $\epsilon\!\sim\!\mathcal{N}(0,\mathbf{I})$ and a scheduler with $(\alpha_t,\sigma_t)$, the noisy latent is
	\begin{equation}
		z_t \;=\; \alpha_t\, z_0 \;+\; \sigma_t\, \epsilon .
	\end{equation}
	A U-Net~\cite{unet} $\epsilon_\theta$ predicts the added noise at step $t$, and the training objective of LDM is
	\begin{equation}
		\mathcal{L}_{\mathrm{LDM}}
		\;=\;
		\mathbb{E}_{z_0=\mathcal{E}(x),\,\epsilon\sim\mathcal{N}(0,\mathbf{I}),\,t}
		\Big[\,
		\|\epsilon - \epsilon_\theta(z_t, t)\|_2^2
		\,\Big].
		\label{eq:ldm}
	\end{equation}
	At inference, the reverse process starts from $z_T\!\sim\!\mathcal{N}(0,\mathbf{I})$  and gradually denoises to $z_0$, followed by $x'=\mathcal{D}(z_0)$.
	
	\noindent\textbf{Cross-Attention in U-Net.} Let $E \in \mathbb{R}^{Z \times d}$ be the text token embeddings (with $Z$ tokens) 
	and $F^{\,l} \in \mathbb{R}^{N \times d}$ be the flattened image features at layer $l$ 
	(with $N=r^2$ image tokens).  
	For a cross-attention in U-Net, the projections are
	\begin{equation}
		Q^{l} = F^{\,l} W_q^{(l)}, 
		\qquad 
		K^{l} = E\, W_k^{(l)},
		\label{eq:qk}
	\end{equation}
	and the cross-attention map is
	\begin{equation}
		A^{l} = \operatorname{softmax}\!\left(\frac{Q^{l}(K^{l})^\top}{\sqrt{d_h}}\right)
		\in \mathbb{R}^{N \times Z},
		\label{eq:attn}
	\end{equation}
	where $A^{l} \in \mathbb{R}^{N\times Z}$ represents the attention weights from all $N$ image tokens to the $Z$ text tokens.

	\noindent\textbf{Latent inpainting.}
	For localized edits and background preservation, inference often blends a proposed edited latent $\tilde z_{t-1}$ with a source latent $z^{\mathrm{src}}_{t-1}$ using a spatial mask $M\!\in\![0,1]^{H_z\times W_z}$ at the latent resolution:
	\begin{equation}
		z_{t-1}
		\;=\;
		M \odot \tilde z_{t-1} \;+\; (1-M) \odot z^{\mathrm{src}}_{t-1}.
		\label{eq:inpaint}
	\end{equation}
	This confines changes to the masked region while keeping the outside region consistent with the source.
	
	\subsection{Delta-Denoising Localization}
	\label{subsec:delta_localization}
	\noindent\textbf{Normal reference guided initialization.}
	To ensure that the synthesized defective images do not deviate completely from the distribution of the original normal samples, we do not start the generation from pure Gaussian noise.
	We encode the guidance image $x^{\mathrm{normal}}$ to $z^{\mathrm{normal}}_{0}$ and partially noise it to an intermediate level $t_{\mathrm{start}}$:
	\begin{equation}
		z^{\mathrm{normal}}_{t_{\mathrm{start}}}
		= \sqrt{\bar{\alpha}_{t_{\mathrm{start}}}}\, z^{\mathrm{normal}}_{0}
		+ \sqrt{1 - \bar{\alpha}_{t_{\mathrm{start}}}}\, \epsilon,
	\end{equation}
	where $t_{\mathrm{start}}{=}\gamma T$ and $\epsilon\sim\mathcal{N}(0,I)$.
	Sampling then starts from $t_{\mathrm{start}}$ and performs the remaining reverse steps ($t_{\mathrm{start}}\!\rightarrow\!0$) for both branches under the same schedule.
	This warm start preserves geometry and illumination, reduces computational cost, and prevents the class-typical drift observed when starting from pure noise.
	
	\noindent\textbf{Prompt design.}
	We use a minimal pair of prompts that differ only by an anomalous attribute. 
	For example, the \emph{normal prompt} is
	\texttt{``a photo of a \{object\}''}, and the \emph{anomaly prompt} is
	\texttt{``a photo of a \{object\} with \{anomaly type\}''}. 
	Additionally, a \emph{description prompt} is introduced for the anomaly token, which provides a refined semantic embedding (e.g., adding shape or intensity cues) during the subsequent prompt refinement stage.  
	This descriptor is not inserted into the sentence but distilled into the anomaly token to enhance its representation to minimize off-target semantic drift.
	
	\noindent\textbf{Synchronized denoising and delta-attribution.}
	Let $T$ be the total number of inference steps. We define $t\!\in\![T,\,T/2]$ as the early denoising stage and $t\!\in\![T/2,\,0]$ as the late denoising stage.
	We run two synchronized trajectories under the same schedule and guidance: a normal branch with prompt $p_{\mathrm n}$ and an anomaly branch with $p_{\mathrm a}$. 
	Denote their latents at step $t$ by $z_t^{(\mathrm n)}$ and $z_t^{(\mathrm a)}\!\in\!\mathbb{R}^{H_z\times W_z\times C}$.
	A single reverse step of the U-Net is
	\begin{equation}
		z_{t-1}^{(\cdot)} \;=\;
		\Phi\!\big(z_t^{(\cdot)},\, \hat{\epsilon}_{\mathrm{cfg}}(z_t^{(\cdot)}, t, e_{(\cdot)}),\, t\big),
		\label{eq:rev}
	\end{equation}
	where $\Phi$ is the reverse-step operator, $\hat{\epsilon}_{\mathrm{cfg}}$ is the classifier-free guided prediction, and $e_{(\cdot)}$ is the text embedding of $p_{\mathrm n}$ or $p_{\mathrm a}$.
	We measure the per-step discrepancy between the two branches after each reverse update:
	\begin{equation}
		d_{t-1}(u) \;=\; \big\|\, z_{t-1}^{(\mathrm n)}(u) - z_{t-1}^{(\mathrm a)}(u) \,\big\|_2 ,
		\label{eq:dt}
	\end{equation}
	where $u$ indexes a spatial position in the latent feature map, and the $\ell_2$ norm is applied over channels.
	The accumulator integrates the discrepancies along the reverse trajectory as
	\begin{equation}
		S_{t-1}(u) \;=\; S_t(u) + d_{t-1}(u), \qquad S_{t_{\mathrm{start}}}(u)=0,
		\label{eq:acc}
	\end{equation}
	where $t_{\mathrm{start}}$ denotes the starting timestep. 
	At the $t_{\mathrm{mid}}{=}\lfloor T/2 \rfloor$, we lightly smooth and normalize $S_{t_{\mathrm{mid}}}$ to obtain a continuous delta map $\hat S$, and threshold it to generate a binary mask:
	\begin{equation}
		M_{\mathrm{mid}}(u) \;=\; \mathbf{1}\!\left\{\, \hat S(u) > \tau_{\mathrm{mid}} \,\right\}.
		\label{eq:mask}
	\end{equation}
	After extracting $M_{\mathrm{mid}}$, the accumulator S is reset to zero, and a new aggregation begins for the late denoising stage, focusing on boundary refinement and local details.
	
	During the late denoising stage, the image-specific mask $M_{\mathrm{mid}}$ serves as a latent inpainting mask that confines subsequent edits to the coarse anomaly region discovered earlier. The update is
	\begin{equation}
		z_{t-1} \leftarrow
		M_{\mathrm{mid}} \odot \tilde z_{t-1} + (1 - M_{\mathrm{mid}})\odot z^{\mathrm{src}}_{t-1},
		\label{eq:inpaint_final}
	\end{equation}
	where the $z^{\mathrm{src}}_{t-1}$ is obtained by forward noising the normal reference latent $z^{\mathrm{normal}}_0$ to the same timestep via DDIM~\cite{ddim}. 
	This operation refines defect synthesis inside the mask while preserving the normal-image distribution outside, ensuring local detail enhancement and global consistency.

	\subsection{Prompt Refinement and Attention Biasing}
	\label{subsec:prompt_attention}
	
	Diffusion-based generation in Stable Diffusion is tightly coupled with the text condition, 
	where each token dynamically interacts with visual features throughout the denoising process. 
	Given our delta-denoising formulation, the semantic discrepancy between the two diffusion branches 
	should be localized exclusively to the anomaly token, 
	rather than spreading across the entire prompt. 
	Therefore, we introduce two complementary mechanisms, \emph{Token-level prompt refinement} and \emph{Spatially-biased cross-attention}.
	
	\noindent\textbf{Token-level prompt refinement.}
	To convey fine-grained defect semantics while minimizing semantic differences outside the anomaly token,
	we perform a lightweight token-level refinement on the anomaly prompt embedding before denoising. 
	Let the normal and anomaly prompts be tokenized as 
	$E_{\mathrm{n}}=\{e_{\mathrm{n}}^1,\ldots,e_{\mathrm{n}}^{Z_{\mathrm{n}}}\}$ and 
	$E_{\mathrm{a}}=\{e_{\mathrm{a}}^1,\ldots,e_{\mathrm{a}}^{Z_{\mathrm{a}}}\}$, respectively. 
	We selectively update $E_{\mathrm{a}}$ with two complementary objectives.
	
	\emph{(i) Anomaly Semantic Distillation.}
	Embeddings of anomaly-specific tokens (e.g., \textit{crack}, \textit{scratch}, \textit{damage}) 
	are steered toward a detailed semantic anchor $e_{\text{detail}}$ distilled from a more descriptive phrase:
	\[
	\mathcal{L}_{\text{anom}} =
	1 - \cos(e_{\mathrm{a}}^j, e_{\text{detail}}) + 
	\lambda \| e_{\mathrm{a}}^j - e_{\text{detail}} \|_2^2 .
	\]
	
	\emph{(ii) Context Alignment for non-anomaly tokens.}
	For the remaining $Z'$ non-anomaly tokens in the anomaly prompt, 
	we compute their mean embedding as a context centroid and encourage each token to stay close to it:
	\[
	\bar e_{\text{ctx}} = \frac{1}{Z'}\sum_{j=1}^{Z'} e_{\mathrm{a}}^{\,j}, \qquad
	\mathcal{L}_{\text{ctx}}
	= \frac{1}{Z'}\sum_{j=1}^{Z'}\| e_{\mathrm{a}}^{\,j} - \bar e_{\text{ctx}} \|_2^2 .
	\]
	This simple MSE regularization stabilizes the shared context and prevents semantic drift among non-anomaly tokens.
	We jointly optimize the two objectives as
	\[
	\mathcal{L}_{\text{prompt}} = \mathcal{L}_{\text{anom}} + \mathcal{L}_{\text{ctx}},
	\]
	and update the anomaly prompt embedding as
	\begin{equation}
		E_{\mathrm{a}} \;\leftarrow\;
		E_{\mathrm{a}} - \eta\, \nabla_{E_{\mathrm{a}}}\mathcal{L}_{\text{prompt}},
		\label{eq:prompt_update}
	\end{equation}
	where $\eta\,$ is the step size.
	This training-free refinement runs for a few iterations with a fixed step size,
	ensuring that semantic changes remain localized to the anomaly token while other tokens retain the original context.

	\noindent\textbf{Spatially-biased cross-attention.}
	We use a two-stage spatial prior for attention. During the early denoising stage, a coarse foreground mask $M_{\mathrm{fg}}$ (from SAM~\cite{sam}) activates plausible surface regions. During the late denoising stage, we switch to the  $M_{\mathrm{mid}}$ from Sec.~\ref{subsec:delta_localization} to focus the defect area.
	At each step, we inject the mask $M$ into the cross-attention logits for the anomaly token only:
	\begin{equation}
		A^{l}
		= \operatorname{softmax}\!\left(
		\frac{\,Q^{l}(K^{l})^{\!\top} + \beta\, M\, o_{a}^{\!\top}\,}{\sqrt{d_h}}
		\right),
		\label{eq:attn_bias}
	\end{equation}
	Here $M$ represents the current spatial prior mask, either $M_{\mathrm{fg}}$ in the early denoising stage or $M_{\mathrm{mid}}$ in the late denoising stage, 
	and $o_a$ denotes the one-hot column selector that specifies the anomaly token $j{=}a$. 
	The bias term $\beta\, M\,o_a^{\!\top}$ injects spatial awareness into the anomaly token’s attention map, 
	so that the model focuses on masked regions when attending to the anomaly description, 
	while leaving other tokens unaffected to maintain overall consistency.
	
	\begin{algorithm}[t]
		\caption{DeltaDeno Inference}
		\label{alg:zsag}
		\begin{algorithmic}[1]
			\STATE $M_{\mathrm{fg}} \!\gets\! \text{SAM}(I)$,\quad $z_0 \!\gets\! \text{VAE-enc}(I)$,\quad $z_t \!\gets\! \text{normal reference guided}(z_0)$
			\STATE Token\mbox{-}level prompt refinement for  $p_{\mathrm a}$
			\STATE Run two diffusion branches with same schedule and seed: $(p_{\mathrm n}, p_{\mathrm a})$
			\FOR{each step $t$}
			\IF{early denoising stage} 
			\STATE attention biasing with $M_{\mathrm{fg}}$; update $S$
			\ELSIF{$t = t_{\mathrm{mid}}$} 
			\STATE $M_{\mathrm{mid}} \!\gets\! \text{threshold(clean(normalize(S)))}$; reset $S$
			\ELSE 
			\STATE attention biasing and latent inpainting with $M_{\mathrm{mid}}$; update $S$
			\ENDIF
			\ENDFOR
			\STATE $M \!\gets\! \text{threshold(clean(normalize(S)))}$,\quad $I' \!\gets\! \text{VAE-dec}(\text{anomaly branch})$
			\STATE \textbf{return} $(I', M)$
		\end{algorithmic}
	\end{algorithm}
	\subsection{Inference}
	
	At inference, ZSAG generates a realistic defect image and its mask from a single normal input.
	Given a normal image $I$, a normal prompt $p_{\mathrm n}$, and an anomaly prompt $p_{\mathrm a}$, we:
	(i) run Token\mbox{-}level prompt refinement for  $p_{\mathrm a}$
	(ii) encode $I$ into latent $z_0$ and get normal reference guided initialization to two synchronized diffusion branches conditioned on $p_{\mathrm n}$ and $p_{\mathrm a}$;
	(iii) accumulate latent discrepancies S during the early denoising stage to form $M_{\mathrm{mid}}$ and reset S at $t_{\mathrm{mid}}$;
	(iv) continue the late denoising stage with attention biasing and inpainting guided by $M_{\mathrm{mid}}$;
	(v) at the final step, re-threshold S to get the final mask $M$, and decode the anomaly branch as the defect image $I'$.

	\section{Experiments}
	\label{sec:exp}
	\subsection{Experiment Setup}
	
	\noindent\textbf{Datasets and Baselines.}
	Following the latest representative ZSAG and FSAG methods~\cite{seas,dual,defectfill}, 
	we mainly conduct experiments on the widely used MVTec AD~\cite{bergmann2019mvtec} and VisA~\cite{visazou2022spot}, which provide a comprehensive benchmark for both generation and detection. 
	We compare with some existing few-shot and zero-shot methods including DRAEM~\cite{draem}, Crop-Paste~\cite{croppaste}, DFMGAN~\cite{dfmgan}, 
	AnomalyDiffusion~\cite{anomalydiffusion}, 
	AnoGen~\cite{anogen},  AnomalyAny~\cite{anomalyany} and AnoStyler~\cite{anostyler}.
	
	\noindent\textbf{Evaluation Metrics.}
	To quantitatively evaluate the anomaly generation quality, 
	we employ the Inception Score (IS) to assess image realism and 
	the intra-cluster pairwise LPIPS distance (IC-LPIPS) to measure generation diversity.
	Following previous FSAG and ZSAG works, we also validate the usefulness of our generated anomalies by testing their effectiveness in anomaly detection frameworks. 
	For anomaly localization and detection performance, 
	we report both image-level and pixel-level metrics, 
	including the Area Under the Receiver Operating Characteristic (AUROC) and 
	the Average Precision (AP). 
	These metrics jointly reflect the localization precision and global detection quality of generated anomalies under a consistent evaluation protocol. 
	
	\noindent\textbf{Implementation Details.}
	Unless otherwise specified, we adopt the pretrained Stable Diffusion v1.5 model as our default backbone, with $T=100$ denoising steps and a partial-noise ratio of $\gamma=0.3$ for a fair comparison with AnomalyAny.
	During generation, we extract the  mask at  $t_{\mathrm{mid}}$ using a threshold of $0.6$, 
	and the final mask at the last step with a threshold of $0.35$ by default. 
	Unless otherwise stated, all other parameters and implementation settings follow the standard Stable Diffusion configuration. 
	As for the  prompt design, we use simple, dataset-agnostic prompt templates during downstream generation. The normal prompt is
	\texttt{``a photo of a \{object\}''}, while the anomaly prompt extends this to \texttt{``a photo of a \{object\} with \{anomaly type\} on it''}.
	For token-level refinement, we additionally use the description prompt in the simple form of
	\texttt{``some large and deep \{anomaly type\}''} or similar variants.

	\begin{figure}[t]
		\centering
		\includegraphics[width=\linewidth]{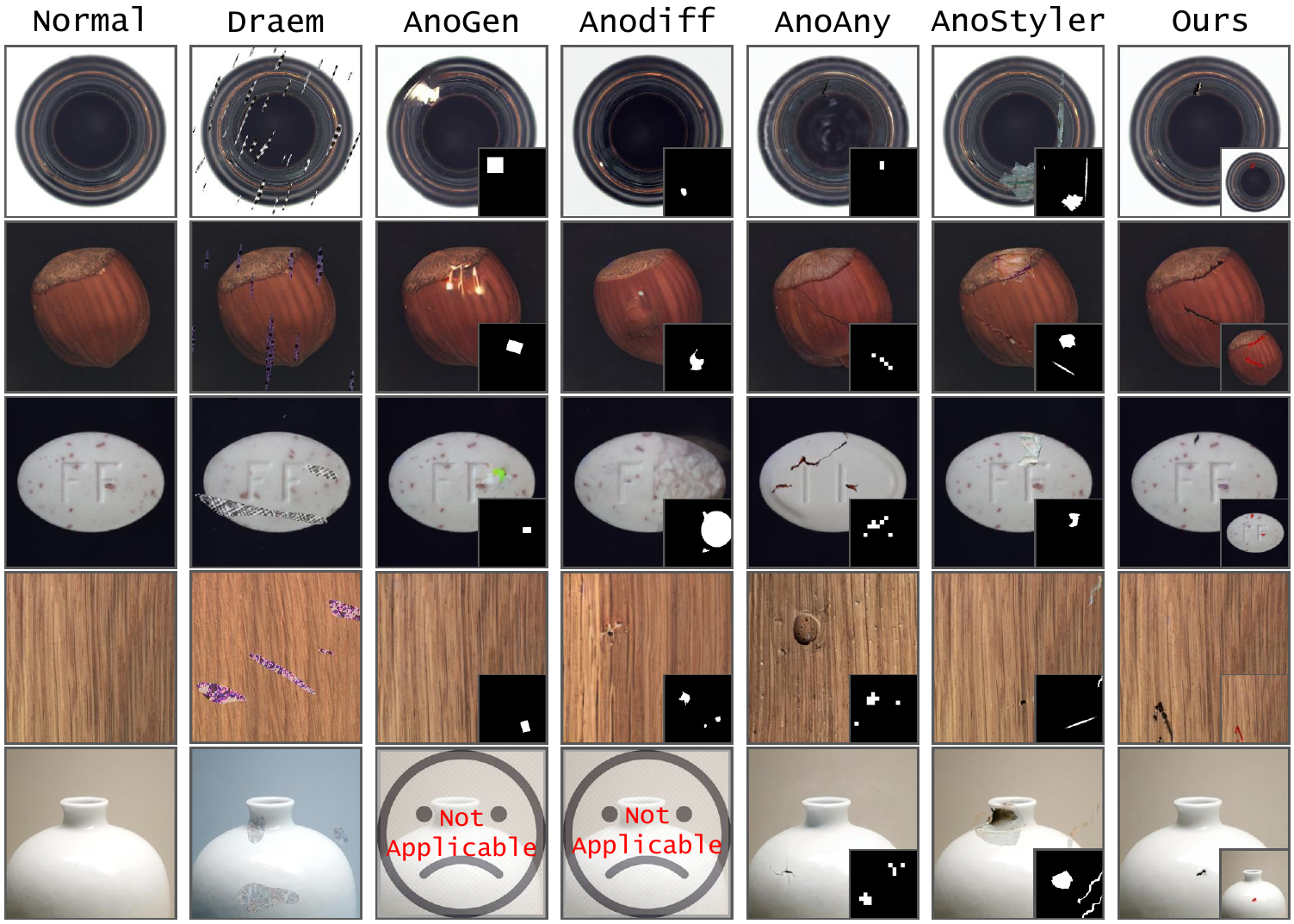}
		\caption{\textbf{Qualitative comparison with existing anomaly generation methods.} Rows 1–4 correspond to categories from MVTec AD, while the last row shows ceramic bottle images collected from the Internet.}
		\label{fig:vis_sompare}
	\end{figure}
	
	\begin{figure}[t]
		\centering
		\begin{subfigure}[t]{0.49\linewidth}
			\centering
			\includegraphics[width=\linewidth]{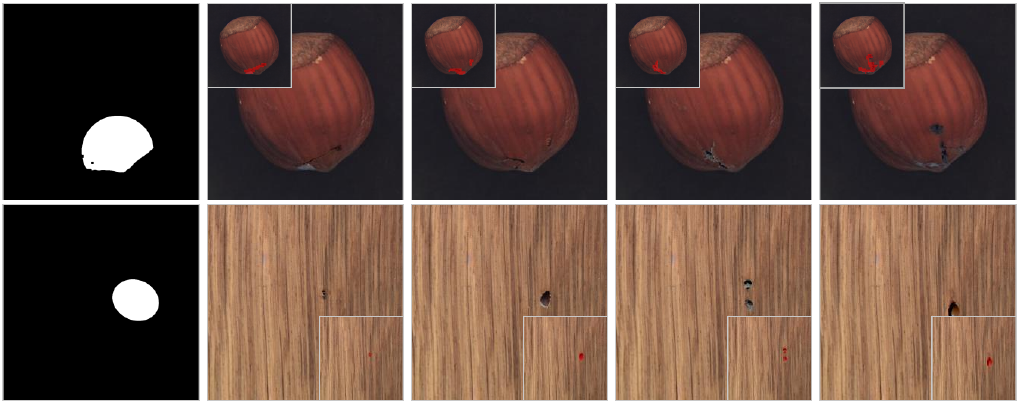}
			\caption{}
			\label{fig:location_control}
		\end{subfigure}
		\hfill
		\begin{subfigure}[t]{0.49\linewidth}
			\centering
			\includegraphics[width=\linewidth]{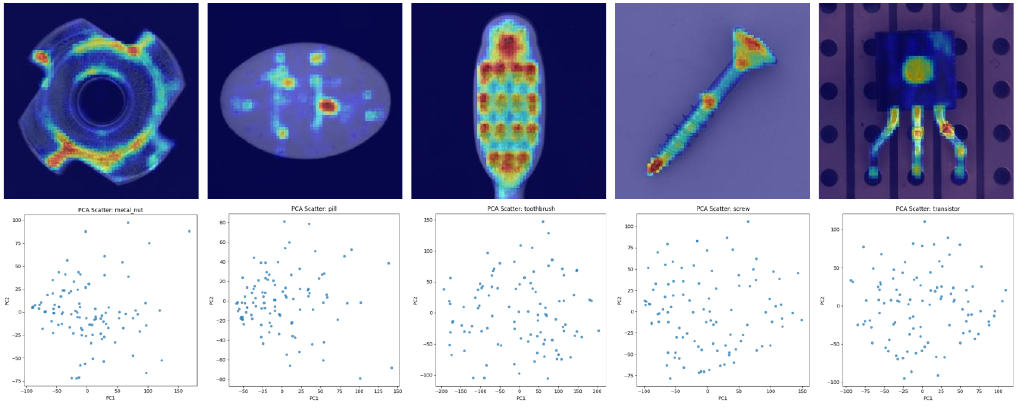}
			\caption{}
			\label{fig:mask_diver}
		\end{subfigure}
		\caption{\textbf{(a)} Illustration of basic spatial location control. 
			The leftmost column shows the manually specified control masks. 
			\textbf{(b)} Spatial frequency maps and PCA visualizations of defect masks generated across multiple categories.}
		\label{fig:two_row_vis}
	\end{figure}
	
	\begin{table}[t]
		\centering
		\caption{\textbf{Comparison with existing anomaly generation methods on generation quality and diversity.}}
		\scriptsize
		\setlength{\tabcolsep}{2pt}
		\renewcommand{\arraystretch}{0.95}
		\resizebox{\textwidth}{!}{
			\begin{tabular}{l|cc|cc|cc|cc|cc|cc}
				\toprule
				Category 
				& \multicolumn{2}{c|}{Crop-Paste~\cite{croppaste}}
				& \multicolumn{2}{c|}{DFMGAN~\cite{dfmgan}}
				& \multicolumn{2}{c|}{AnomalyDiff~\cite{anomalydiffusion}}
				& \multicolumn{2}{c|}{AnomalyAny~\cite{anomalyany}}
				& \multicolumn{2}{c|}{AnoStyler~\cite{anostyler}}
				& \multicolumn{2}{c}{Ours} \\
				& IS↑ & IC-L↑ 
				& IS↑ & IC-L↑ 
				& IS↑ & IC-L↑ 
				& IS↑ & IC-L↑ 
				& IS↑ & IC-L↑
				& IS↑ & IC-L↑ \\
				\midrule
				bottle       & 1.43 & 0.04 & 1.62 & 0.12 & 1.58 & 0.19 & 1.73 & 0.17 & 1.55 & 0.21 & 2.36 & 0.27 \\
				cable        & 1.74 & 0.25 & 1.96 & 0.25 & 2.13 & 0.41 & 2.06 & 0.41 & 1.53 & 0.39 & 2.49 & 0.41 \\
				capsule      & 1.23 & 0.05 & 1.59 & 0.11 & 1.59 & 0.21 & 2.16 & 0.23 & 1.96 & 0.20 & 1.82 & 0.23 \\
				carpet       & 1.17 & 0.11 & 1.23 & 0.13 & 1.16 & 0.24 & 1.10 & 0.34 & 1.28 & 0.26 & 1.27 & 0.45 \\
				grid         & 2.00 & 0.12 & 1.97 & 0.13 & 2.04 & 0.44 & 2.31 & 0.38 & 2.52 & 0.39 & 2.26 & 0.39 \\
				hazelnut     & 1.74 & 0.21 & 1.93 & 0.24 & 2.13 & 0.31 & 2.55 & 0.32 & 1.83 & 0.32 & 2.62 & 0.38 \\
				leather      & 1.47 & 0.14 & 2.06 & 0.17 & 1.94 & 0.41 & 2.26 & 0.41 & 2.94 & 0.45 & 1.77 & 0.44 \\
				metal\_nut   & 1.56 & 0.15 & 1.49 & 0.32 & 1.96 & 0.30 & 1.82 & 0.27 & 2.33 & 0.30 & 2.00 & 0.31 \\
				pill         & 1.49 & 0.11 & 1.63 & 0.16 & 1.61 & 0.26 & 2.91 & 0.30 & 1.65 & 0.24 & 2.26 & 0.33 \\
				screw        & 1.12 & 0.16 & 1.12 & 0.14 & 1.28 & 0.30 & 1.33 & 0.32 & 1.39 & 0.32 & 1.29 & 0.37 \\
				tile         & 1.83 & 0.20 & 2.39 & 0.22 & 2.54 & 0.55 & 2.66 & 0.53 & 2.92 & 0.52 & 2.30 & 0.50 \\
				toothbrush   & 1.30 & 0.08 & 1.82 & 0.18 & 1.68 & 0.21 & 1.64 & 0.22 & 1.46 & 0.17 & 1.99 & 0.24 \\
				transistor   & 1.39 & 0.15 & 1.64 & 0.25 & 1.57 & 0.34 & 1.66 & 0.28 & 1.57 & 0.29 & 1.97 & 0.32 \\
				wood         & 1.95 & 0.23 & 2.12 & 0.35 & 2.33 & 0.37 & 1.93 & 0.41 & 2.91 & 0.39 & 2.19 & 0.44 \\
				zipper       & 1.23 & 0.11 & 1.29 & 0.27 & 1.39 & 0.25 & 2.14 & 0.33 & 2.77 & 0.29 & 2.09 & 0.37 \\
				\midrule
				\textbf{Avg.} 
				& 1.51 & 0.14 
				& 1.72 & 0.20 
				& 1.80 & 0.32 
				& 2.02 & 0.33 
				& 2.04 & 0.32 
				& \textbf{2.05} & \textbf{0.36} \\
				\bottomrule
		\end{tabular}}
		\label{tab:metric1}
	\end{table}
	
	\begin{table}[t]
		\centering
		\caption{\textbf{Comparison of downstream detection performance on MVTec AD from ZSAG methods}. Bold numbers indicate the best results.}
		\label{tab:mvtec}
		\scriptsize
		\setlength{\tabcolsep}{1.2pt}
		\renewcommand{\arraystretch}{0.95}
		\resizebox{\linewidth}{!}{
			\begin{tabular}{l|cccc|cccc|cccc|cccc}
				\toprule
				& \multicolumn{4}{c|}{AnomalyAny (CVPR'25)} & \multicolumn{4}{c|}{AnoStyler (AAAI'26)} & \multicolumn{4}{c|}{DeltaDeno (SD1.5)} & \multicolumn{4}{c}{DeltaDeno (FLUX)} \\
				\cmidrule(lr){2-5} \cmidrule(lr){6-9} \cmidrule(lr){10-13} \cmidrule(lr){14-17}
				Category 
				& AUC$_i$ & AP$_i$ & AUC$_p$ & AP$_p$
				& AUC$_i$ & AP$_i$ & AUC$_p$ & AP$_p$
				& AUC$_i$ & AP$_i$ & AUC$_p$ & AP$_p$ 
				& AUC$_i$ & AP$_i$ & AUC$_p$ & AP$_p$ \\
				\midrule
				bottle      & 94.9 & 98.0 & 86.9 & 32.2 & 92.3 & 97.3 & 67.6 & 17.7 & \textbf{97.2} & \textbf{98.8} & 86.0 & 28.4 & 92.7 & 97.1 & \textbf{91.1} & \textbf{45.1} \\
				cable       & 44.1 & 49.0 & 82.2 & 11.1 & 62.0 & 63.3 & 65.3 & 4.4 & 70.9 & 74.9 & 76.2 & 8.0 & \textbf{89.3} & \textbf{91.7} & \textbf{86.6} & \textbf{24.5} \\
				capsule     & 59.4 & 84.6 & 78.2 & 7.4 & 80.9 & \textbf{93.6} & 84.8 & 13.0 & 54.9 & 79.0 & \textbf{87.9} & 5.8 & \textbf{82.9} & 93.1 & 78.4 & \textbf{18.4} \\
				carpet      & 79.4 & 90.2 & 80.9 & 30.9 & 81.7 & 92.7 & 90.1 & 36.2 & 85.9 & 94.0 & 90.1 & 41.7 & \textbf{89.1} & \textbf{95.3} & \textbf{95.8} & \textbf{61.8} \\
				grid        & 85.7 & 92.7 & 73.9 & 5.3 & 71.0 & 83.0 & 55.2 & 0.9 & 81.3 & 89.9 & 77.3 & 8.7 & \textbf{86.5} & \textbf{93.5} & \textbf{87.4} & \textbf{24.3} \\
				hazelnut    & 98.8 & 99.0 & 95.3 & 43.8 & 97.6 & 98.8 & 97.5 & 71.2 & 99.1 & 99.2 & 97.5 & 69.6 & \textbf{99.2} & \textbf{99.4} & \textbf{99.1} & \textbf{77.5} \\
				leather     & 92.6 & 97.0 & 94.1 & 45.0 & \textbf{98.6} & \textbf{99.8} & 94.2 & 36.4 & 96.5 & 98.3 & 94.6 & \textbf{47.1} & 86.2 & 94.1 & \textbf{97.2} & 45.1 \\
				metal\_nut  & 54.3 & 80.9 & 77.1 & 41.0 & 65.8 & 87.6 & 56.7 & 23.7 & 73.8 & 90.0 & 83.2 & 38.5 & \textbf{85.6} & \textbf{95.0} & \textbf{92.2} & \textbf{69.1} \\
				pill        & 65.9 & 88.9 & 88.9 & 36.7 & 82.3 & 95.1 & 84.2 & 59.4 & \textbf{89.7} & \textbf{97.3} & 93.3 & \textbf{65.1} & 87.7 & 96.4 & \textbf{95.4} & \textbf{65.1} \\
				screw       & 60.9 & 75.7 & 92.1 & 2.3 & 84.9 & 93.2 & 93.2 & \textbf{15.9} & 73.3 & 85.8 & \textbf{95.6} & 4.7 & \textbf{100.0} & \textbf{100.0} & 46.6 & 0.2 \\
				tile        & 97.0 & 98.5 & 89.2 & 53.9 & 99.1 & 99.5 & 94.2 & 75.8 & 99.9 & \textbf{100.0} & 87.0 & 51.9 & \textbf{100.0} & \textbf{100.0} & \textbf{98.7} & \textbf{88.3} \\
				toothbrush  & 89.6 & 95.0 & 75.2 & 12.2 & 87.8 & 92.3 & 77.8 & 10.8 & 89.2 & 94.1 & 78.9 & 12.9 & \textbf{90.0} & \textbf{95.3} & \textbf{86.6} & \textbf{19.9} \\
				transistor  & 78.9 & 72.4 & \textbf{65.2} & 12.0 & \textbf{84.6} & \textbf{84.4} & 58.4 & \textbf{17.0} & 78.8 & 76.8 & 58.9 & 14.9 & \textbf{84.6} & 81.3 & 62.8 & \textbf{17.0} \\
				wood        & \textbf{99.4} & \textbf{99.7} & 84.4 & 42.1 & 94.4 & 98.0 & \textbf{84.9} & \textbf{56.4} & 99.0 & 99.6 & 80.7 & 42.2 & 96.9 & 98.6 & 81.7 & 44.0 \\
				zipper      & 63.6 & 84.8 & 85.7 & 21.8 & \textbf{83.5} & \textbf{94.7} & 68.4 & 14.8 & 80.8 & 91.7 & \textbf{86.8} & 18.5 & 74.9 & 90.7 & 81.8 & \textbf{22.5} \\
				\midrule
				\textbf{Mean} & 77.6 & 87.1 & 83.3 & 26.5 & 84.4 & 91.5 & 78.2 & 30.2 & 84.7 & 91.3 & 84.9 & 30.5 & \textbf{89.7} & \textbf{94.8} & \textbf{85.4} & \textbf{41.5} \\
				\bottomrule
		\end{tabular}}
	\end{table}
	
	\begin{table}[t]
		\centering
		\caption{\textbf{Comparison of downstream detection performance on VisA from ZSAG methods}. Bold numbers indicate the best results.}
		\label{tab:visa}
		\scriptsize
		\setlength{\tabcolsep}{1.2pt}
		\renewcommand{\arraystretch}{0.95}
		\resizebox{\linewidth}{!}{
			\begin{tabular}{l|cccc|cccc|cccc|cccc}
				\toprule
				& \multicolumn{4}{c|}{AnomalyAny (CVPR'25)} & \multicolumn{4}{c|}{AnoStyler (AAAI'26)} & \multicolumn{4}{c|}{DeltaDeno (SD1.5)} & \multicolumn{4}{c}{DeltaDeno (FLUX)} \\
				\cmidrule(lr){2-5} \cmidrule(lr){6-9} \cmidrule(lr){10-13} \cmidrule(lr){14-17}
				Category 
				& AUC$_i$ & AP$_i$ & AUC$_p$ & AP$_p$
				& AUC$_i$ & AP$_i$ & AUC$_p$ & AP$_p$
				& AUC$_i$ & AP$_i$ & AUC$_p$ & AP$_p$ 
				& AUC$_i$ & AP$_i$ & AUC$_p$ & AP$_p$ \\
				\midrule
				candle      & 50.9 & 56.7 & \textbf{87.8} & 0.8 & 71.2 & 66.2 & 80.8 & \textbf{16.5} & 60.0 & 67.4 & 85.9 & 6.9 & \textbf{81.9} & \textbf{82.7} & 78.9 & 1.0 \\
				capsules    & 81.3 & 86.6 & 76.8 & 2.7 & 62.5 & 75.8 & 64.9 & 0.7 & \textbf{91.6} & \textbf{94.1} & 92.0 & 13.3 & 79.5 & 87.4 & \textbf{92.2} & \textbf{21.5} \\
				cashew      & 82.6 & 85.6 & \textbf{90.1} & 6.2 & 85.3 & 87.4 & \textbf{90.1} & \textbf{7.8} & 88.8 & 92.6 & 78.3 & 4.4 & \textbf{95.8} & \textbf{97.7} & 66.7 & 2.0 \\
				chewinggum  & 89.3 & 95.4 & 94.9 & 44.9 & 94.6 & 97.5 & 97.6 & \textbf{71.1} & \textbf{95.4} & \textbf{97.9} & 97.3 & 37.9 & 91.5 & 96.3 & \textbf{98.0} & 49.4 \\
				fryum       & 86.6 & 93.7 & 86.1 & 11.0 & \textbf{90.2} & \textbf{94.0} & 72.2 & 10.9 & 78.2 & 88.7 & 87.1 & 9.4 & 77.0 & 88.6 & \textbf{92.2} & \textbf{20.6} \\
				macaroni1   & 92.1 & 92.0 & 92.9 & 0.5 & 91.4 & 91.5 & 97.1 & \textbf{25.0} & \textbf{94.8} & \textbf{94.9} & \textbf{98.3} & 16.6 & 87.8 & 85.4 & 93.5 & 0.4 \\
				macaroni2   & 46.2 & 46.5 & 94.4 & 0.4 & 67.3 & \textbf{69.7} & \textbf{97.4} & \textbf{20.6} & \textbf{70.5} & 68.9 & 96.9 & 3.4 & 65.9 & 67.3 & 94.2 & 2.8 \\
				pcb1        & 69.4 & 69.6 & 90.9 & 3.1 & 49.2 & 48.3 & 73.2 & 0.8 & 72.3 & 77.4 & 86.5 & \textbf{7.8} & \textbf{89.0} & \textbf{85.3} & \textbf{95.3} & 6.2 \\
				pcb2        & 76.5 & 83.9 & 82.9 & 0.9 & 67.6 & 78.2 & 88.5 & 1.8 & \textbf{85.6} & \textbf{86.1} & \textbf{90.7} & \textbf{9.1} & 92.6 & 93.7 & 89.1 & 5.5 \\
				pcb3        & 49.9 & 49.9 & 84.7 & 1.1 & 81.5 & 84.4 & 84.0 & \textbf{10.7} & 72.1 & 76.5 & \textbf{88.5} & 2.2 & \textbf{83.2} & \textbf{85.6} & 87.3 & 7.9 \\
				pcb4        & 87.1 & 82.7 & 86.0 & 2.5 & 78.9 & 86.0 & 89.8 & 11.6 & \textbf{96.6} & \textbf{95.4} & 85.2 & 7.7 & 93.9 & 90.6 & \textbf{95.6} & \textbf{13.3} \\
				pipe\_fryum & \textbf{84.2} & \textbf{92.7} & 83.5 & 16.2 & 79.0 & 89.4 & 96.0 & 22.2 & 79.5 & 88.9 & 94.7 & 20.1 & 82.8 & 90.7 & \textbf{98.9} & \textbf{49.6} \\
				\midrule
				\textbf{Mean} & 74.7 & 77.9 & 87.6 & 7.5 & 76.5 & 80.7 & 86.0 & \textbf{16.6} & 82.1 & 85.7 & 90.1 & 11.6 & \textbf{85.1} & \textbf{87.6} & \textbf{90.2} & 15.0 \\
				\bottomrule
		\end{tabular}}
	\end{table}
	
	\subsection{Anomaly Generation Results}
	
	\noindent\textbf{Qualitative visualization.}
	We conduct a qualitative comparison on major MVTec AD categories and unseen real industrial samples (Fig.~\ref{fig:vis_sompare}). 
	DRAEM often exhibits illumination mismatch and visible boundary seams due to cut-paste composition. 
	FSAG methods (AnomalyDiffusion, AnoGen) rely on few-shot fine-tuning with real defects and masks, performing well on seen categories but failing to generalize to unseen ones. 
	AnomalyAny produces coarse masks derived from low-resolution cross-attention, leading to blurry boundaries and inaccurate localization. 
	AnoStyler sometimes generates defects that are semantically inconsistent with the target category, reducing realism. 
	In contrast, our method produces visually sharp boundaries and context-consistent defects across both seen and unseen categories.
	To further illustrate the generation quality, Fig.~\ref{fig:two_row_vis}(a) demonstrates basic spatial location control, where the generated anomalies follow the specified masks. 
	Fig.~\ref{fig:two_row_vis}(b) presents the spatial frequency maps and PCA projections of generated defect masks across multiple categories, revealing diverse spatial distributions and structural variations.

	\noindent\textbf{Generation quality.}
	We quantitatively evaluate anomaly image quality on MVTec AD using Inception Score (IS) for fidelity and IC\mbox{-}LPIPS for diversity (Tab.~\ref{tab:metric1}). 
	Our method consistently achieves higher IS and IC\mbox{-}LPIPS than prior works, indicating improved photorealistic fidelity while maintaining diverse defect patterns. 
	These results confirm that our approach balances high-quality synthesis with cross-category generalization in a fully training-free setting.
	
	\subsection{Anomaly Detection Results}
	
	We evaluate the effectiveness of the generated anomalous data via a downstream detection task on MVTec AD and VisA. Following AnomalyAny, we generate 100 anomalous images per category from the single normal image  to ensure a fair comparison and train a simple U\mbox{-}Net on the synthesized data. We report performance on the real MVTec AD and VisA  test split (Tab.~\ref{tab:mvtec}, Tab.~\ref{tab:visa}).
	For ZSAG methods,  we further control the source by generating from the same single normal image per category to isolate the generation effect. No anomalous samples from the real datasets are used during our synthesis or U\mbox{-}Net training.  All results are obtained with the same U\mbox{-}Net architecture and training schedule without test\mbox{-}time adaptation.

	Results show that our synthesized data yield competitive or superior downstream detection scores compared to recent ZSAG methods across the majority of tested categories. Overall, these gains stem from jointly balancing localization precision, contextual fidelity, and morphological diversity during synthesis. Crucially, our delta-denoising framework is intrinsically model-agnostic, exhibiting strong transferability and generality. As demonstrated by the seamless transition from the default Stable Diffusion v1.5 to the more advanced FLUX.1-dev~\cite{flux2024} backbone, our method scales effectively and yields further performance boosts without requiring architecture-specific tuning. Nevertheless, under the strict ZSAG setting, the synthesized defect-type distribution inevitably deviates from real-world data. Thus, for categories with highly complex or fine-grained normal patterns, detection gains may remain limited.
	\begin{figure}[t]
		\centering
		\includegraphics[width=0.45\linewidth]{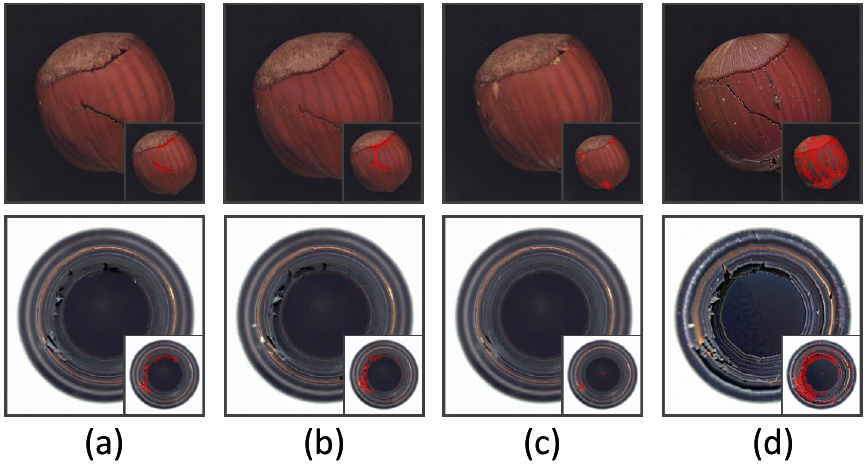}
		\hfill
		\includegraphics[width=0.53\linewidth]{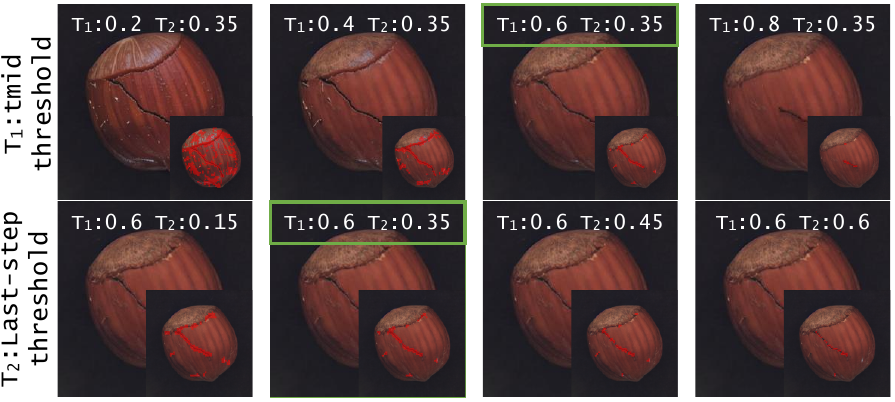}
		\caption{
			\textbf{Ablation study.}
			\textbf{Left:} Module ablation of \emph{DeltaDeno}.
			(a) Full model.
			(b) w/o $L_{\text{ctx}}$.
			(c) w/o spatial attention biasing.
			(d) w/o latent mask inpainting in the late denoising stage.
			\textbf{Right:} Threshold ablation showing the effect of different threshold values on generation results.
		}
		\label{fig:ablation_all}
	\end{figure}
	
	\begin{figure}[t]
		\centering
		\includegraphics[width=\linewidth]{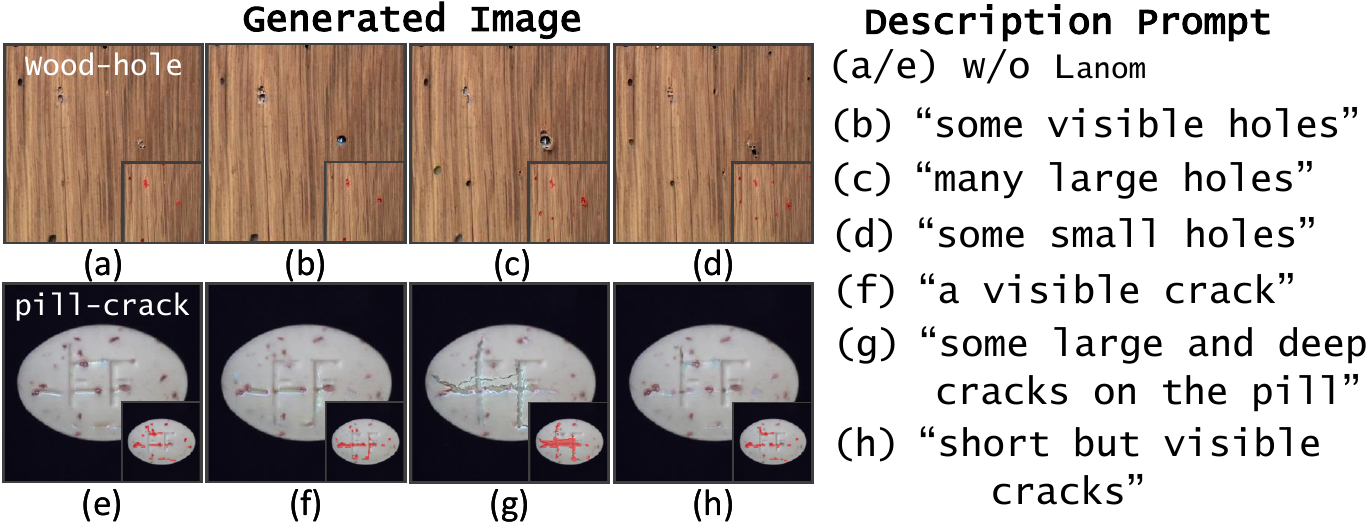}
		\caption{\textbf{Ablation visualization of the Anomaly Semantic Distillation module.} Varying the description prompts steers the generated defect type and severity, yielding more controllable and diverse anomalies. }
		\label{fig:ablation2}
	\end{figure}
	\begin{table}[t]
		\centering
		\caption{\textbf{Ablation on downstream detection (mean \%).} Higher is better.
			\emph{A}: latent-mask inpainting,\; \emph{B}: attention biasing,\; \emph{C}: prompt refinement.}
		\small
		\setlength{\tabcolsep}{4pt}
		\begin{tabular}{lcccc}
			\toprule
			Method & AUC$_i$ $\uparrow$ & AP$_i$ $\uparrow$ & AUC$_p$ $\uparrow$ & AP$_p$ $\uparrow$ \\
			\midrule
			w/o A & 79.7 & 87.6 & 81.2 & 25.8 \\
			w/o B & 79.5 & 88.2 & 83.4 & 28.3 \\
			w/o C & 78.8 & 86.3 & 83.1 & 28.1 \\
			\midrule
			\textbf{DeltaDeno (full)} & \textbf{84.7} & \textbf{91.3} & \textbf{84.9} & \textbf{30.5} \\
			\bottomrule
		\end{tabular}
		\label{tab:ablation3}
	\end{table}

	\subsection{Ablation Studies}
	We conduct both qualitative and quantitative ablations on MVTec AD to analyze the contribution of each component and the sensitivity to key hyperparameters.
	
	\noindent\textbf{Module ablation.}
	We first perform a qualitative module ablation as shown in Fig.~\ref{fig:ablation_all} (left). 
	Without the context alignment term $L_{\text{ctx}}$, non-anomaly tokens are more likely to drift, causing off-target edits that reduce mask localization accuracy. 
	Removing latent-mask inpainting in the late denoising stage leads to noticeable spill-over edits and degraded background consistency, since the generation is no longer confined to the estimated defect region. 
	Without spatial attention biasing, the anomaly-token semantics are not effectively activated in the target region, often leading to weak or even missing defect synthesis.
	Overall, the full model produces the sharpest boundaries and the most coherent defect synthesis.

	\noindent\textbf{Threshold tuning.}
	We further study the sensitivity to the two thresholds used in our DeltaDeno (Fig.~\ref{fig:ablation_all}, right). 
	The $t_{\text{mid}}$ threshold mainly controls the size of the edited anomaly region during the late denoising stage, and thus has a larger influence on the final synthesized defect appearance. 
	In contrast, the last-step threshold primarily affects the precision of the extracted mask, while having negligible impact on the visual quality of the generated image. 
	Since the optimal thresholds vary across categories, we adopt a unified default setting of $0.6$--$0.35$ for all categories in our main experiments based on this observation.
	
	\noindent\textbf{Anomaly semantic distillation.}
	Fig.~\ref{fig:ablation2} visualizes the effect of the Anomaly Semantic Distillation module. 
	By varying the description prompts, the anomaly token is anchored to different fine-grained attributes (e.g., length, width, texture, or severity), which steers the resulting defect morphology accordingly. 
	This provides controllable generation while maintaining contextual consistency around the defect region.
	
	\noindent\textbf{Quantitative ablation.}
	Finally, we evaluate the practical utility of each module via downstream detection (Tab.~\ref{tab:ablation3}). 
	All variants follow the same protocol: for each category we synthesize the same number of anomalies from the same normal guidance image, train a U-Net on the generated data, and evaluate on the MVTec AD test split. 
	The full configuration achieves the best image-level and pixel-level metrics. 
	Removing any component consistently degrades performance, confirming that latent-mask inpainting, attention biasing, and prompt refinement are all beneficial and complementary.

	\section{Conclusion}
	\label{sec:conclusion}
	We introduced \emph{DeltaDeno}, a training-free zero-shot anomaly generation framework that leverages delta-denoising to localize and edit defects in a normal sample. By contrasting two synchronized diffusion branches conditioned on normal and anomaly prompts, the method derives reliable masks and performs mask-guided latent inpainting to synthesize photorealistic, context-consistent anomalies. A normal-reference initialization keeps synthesis aligned with the distribution of normal content, while lightweight prompt refinement with spatially biased attention stabilizes localization without retraining. On public benchmarks, \emph{DeltaDeno} attains higher generation quality and stronger downstream detection performance. The framework requires no anomalous labels or model adaptation and is practical for rapid data bootstrapping and stress testing when real defects are scarce.
	
	\noindent\textbf{Limitations and Future Work.}
	Despite its effectiveness, \emph{DeltaDeno} still faces several limitations. 
	First, existing evaluation protocols for ZSAG are limited and may require MLLM-based comprehensive assessment in the future. 
	Second, the controllability offered by semantic distillation remains limited, constraining fine-grained manipulation of defect attributes. 
	Third, most generated anomalies exhibit structural or surface-level characteristics, while more abstract or logical defects remain challenging to produce. 
	Future work can investigate these unresolved aspects to further improve the capability of zero\mbox{-}shot anomaly generation.

	\section*{Acknowledgements}
	This work was supported in part by the National Science and Technology Major Project under Grant 2022ZD0119400, the National Natural Science Foundation of China under Grant No. 62303458 and 62303461, and the Beijing Municipal Natural Science Foundation (China) under Grant 4252053.

	%
	%

	\bibliographystyle{splncs04}
	\bibliography{main}
	
	\clearpage
	\appendix
	\section*{Appendix}
	
	\section{Overview}
	\label{sec:overview}
	
	This appendix provides additional technical details, extended analyses, and supplementary experiments for the DeltaDeno framework. It complements the main paper by offering a deeper examination of implementation components, a critical discussion of evaluation protocols for zero-shot anomaly generation, and further empirical evidence on the generalization ability of our method.
	
	The appendix is organized as follows:
	\begin{itemize}
		\item \textbf{App.~\ref{app:impl}} presents extended implementation details, including the diffusion backbone configuration, prompt design, and the evaluation metrics and training protocols used throughout the experiments.
		\item \textbf{App.~\ref{app:additional_experiments}} presents additional experiments and ablations, including a quantitative comparison with few-shot anomaly generation (FSAG) methods, an analysis of the method's dependence on SAM, evaluations of runtime and memory efficiency, and a detailed quantitative threshold analysis.
		\item \textbf{App.~\ref{app:discussion}} provides an in-depth discussion of current evaluation practices for anomaly generation, covering generation-quality metrics, downstream detection protocols, and the often-overlooked quality of mask--image pairs.
		\item \textbf{App.~\ref{app:vis}} offers extended qualitative visualizations of synthesized anomalous images and their corresponding masks across multiple industrial categories.
	\end{itemize}

	\section{Extended Implementation Details}
	\label{app:impl}
	
	\subsection{Implementation Details}
	\label{app:impl:backbone}
	We build DeltaDeno on Stable Diffusion v1.5 and keep the original UNet, VAE, and text encoder settings. All experiments use $T=100$ denoising steps and a partial-noise ratio of $\gamma = 0.3$. Delta maps and masks are computed on the latent feature maps at a fixed spatial resolution of $64 \times 64$. 
	For all FLUX-based experiments, we use the DeltaDeno pipeline with 30 denoising steps, a guidance scale of 3.5, and a strength of 0.8. 
	
	During sampling, we extract a mid-level mask at $t_{\mathrm{mid}}$ with a normalized threshold of $0.6$, and a final mask at the last step with a threshold of $0.35$. The spatial prior used for cross-attention biasing is applied with strength $\beta = 4$. For prompt refinement, the anomaly-token loss weight in $\mathcal{L}_{\text{anom}}$ is set to $\lambda = 0.5$, and the anomaly prompt embedding $E_{\mathrm{a}}$ is updated with a fixed step size $\eta = 4$ at each refinement step.
	
	\subsection{Prompt Design}
	\label{app:impl:prompt}
	
	We use simple, dataset-agnostic prompt templates. The normal prompt is
	\texttt{``a photo of a \{object\}''}, where \texttt{\{object\}} is the category name provided by the dataset (e.g., \texttt{bottle}, \texttt{hazelnut}). The anomaly prompt extends this with an explicit defect description:
	\texttt{``a photo of a \{object\} with \{anomaly type\} on it''}, where \texttt{\{anomaly type\}} denotes the defect type (e.g., \texttt{crack}, \texttt{hole}, \texttt{scratch}). 
	
	For token-level refinement we additionally use a more detailed textual phrase, referred to as the description prompt, in the form of
	\texttt{``some large and deep \{anomaly type\}''} or similar variants, which provides a richer semantic target for the anomaly-related tokens while keeping the object description unchanged.
	
	For categories with coarse labels (e.g., \texttt{toothbrush} with only \texttt{defective}), we adopt generic anomaly terms such as \texttt{damage} or \texttt{defect}, optionally with simple modifiers (e.g., \texttt{visible}, \texttt{small}), consistent with our prompt design for other categories. 
	Since we consistently use generic anomaly terms, the downstream detection performance is largely insensitive to specific prompt choices, which we also observe empirically.

	\subsection{Evaluation Metrics and Protocols}
	\label{sec:eval_protocols}
	
	Following established protocols in few-shot anomaly generation literature, such as AnomalyDiffusion~\cite{anomalydiffusion}, we evaluate our proposed DeltaDeno framework from two complementary perspectives: \textit{visual generation quality} and \textit{downstream detection effectiveness}.
	
	\noindent\textbf{Generation Quality and Diversity.}
	To quantitatively assess the visual fidelity and diversity of the synthesized anomalies, we employ two standard metrics: Inception Score (IS) and Intra-Cluster Pairwise LPIPS (IC-LPIPS). For evaluation, we generate 1,000 anomaly samples for each category. We use IS to measure the perceptual quality and realism of the generated images, where a higher score indicates that the synthesized anomalies exhibit clear and distinct features consistent with real-world distributions. In parallel, to evaluate the diversity of samples within each category, we compute IC-LPIPS by averaging the pairwise distances between all generated images; a higher IC-LPIPS score indicates greater variation in defect appearance, suggesting that the model avoids mode collapse and produces diverse anomaly patterns.
	
	\noindent\textbf{Downstream Detection Effectiveness.}
	To validate the practical utility of our generated data, we conduct a downstream anomaly detection task following the one-shot setting in AnomalyAny. For each category, we synthesize 100 anomalous images derived from a single normal reference image to strictly isolate the generation effect. During this generation process, we employ a standard set of generic anomaly types (e.g., \textit{crack}, \textit{hole}, \textit{damage}) and utilize simple descriptive modifiers (e.g., \textit{large}, \textit{deep}, \textit{small}) for the description prompts.
	
	Adhering to the standard one-class anomaly detection setting, we train a dedicated U-Net~\cite{unet} segmentation network for each object category. We follow the single-class AD setting commonly adopted in prior work, as multi-class training typically leads to worse detection performance. These category-specific models are trained from scratch utilizing solely the synthetic samples, without exposure to any real anomalous data. Crucially, to ensure a fair and consistent comparison with few-shot methods (e.g., DFMGAN~\cite{dfmgan}, AnomalyDiffusion~\cite{anomalydiffusion}) that typically utilize the first one-third of real anomalous samples for training, we evaluate our trained detector exclusively on the remaining two-thirds of the MVTec AD test set. We report pixel-level and image-level AUROC and Average Precision (AP) on this held-out split to measure how well the synthetic distribution aligns with unseen real-world defects.
	
	\section{Additional Experiments}
	\label{app:additional_experiments}
	
	\subsection{Comparison with Few-Shot Anomaly Generation Methods}
	
	\begin{table}[h]
		\centering
		\caption{\textbf{Quantitative comparison between FSAG methods and our DeltaDeno on MVTec AD}. Bold numbers indicate the best results across all methods.}
		\label{tab:mvtec_vs_fsag}
		\scriptsize
		\setlength{\tabcolsep}{1.2pt}
		\renewcommand{\arraystretch}{0.95}
		\resizebox{\linewidth}{!}{
			\begin{tabular}{l|cccc|cccc|cccc|cccc}
				\toprule
				& \multicolumn{4}{c|}{AnoDiff (AAAI'24)} & \multicolumn{4}{c|}{AnoGen (ECCV'24)} & \multicolumn{4}{c|}{DeltaDeno (SD1.5)} & \multicolumn{4}{c}{DeltaDeno (FLUX)} \\
				\cmidrule(lr){2-5} \cmidrule(lr){6-9} \cmidrule(lr){10-13} \cmidrule(lr){14-17}
				Category 
				& AUC$_i$ & AP$_i$ & AUC$_p$ & AP$_p$
				& AUC$_i$ & AP$_i$ & AUC$_p$ & AP$_p$
				& AUC$_i$ & AP$_i$ & AUC$_p$ & AP$_p$ 
				& AUC$_i$ & AP$_i$ & AUC$_p$ & AP$_p$ \\
				\midrule
				bottle      & \textbf{97.7} & \textbf{99.3} & \textbf{94.6} & \textbf{81.6} & 96.6 & 98.8 & 75.6 & 57.4 & 97.2 & 98.8 & 86.0 & 28.4 & 92.7 & 97.1 & 91.1 & 45.1 \\
				cable       & \textbf{99.5} & \textbf{99.5} & \textbf{93.2} & \textbf{74.2} & 95.8 & 97.0 & 81.2 & 57.5 & 70.9 & 74.9 & 76.2 & 8.0  & 89.3 & 91.7 & 86.6 & 24.5 \\
				capsule     & 77.4 & 94.3 & 74.8 & 12.6          & 78.9 & 94.0 & 62.9 & \textbf{19.4} & 54.9 & 79.0 & \textbf{87.9} & 5.8  & \textbf{82.9} & 93.1 & 78.4 & 18.4 \\
				carpet      & 64.8 & 79.4 & 79.8 & 17.8          & \textbf{90.0} & \textbf{96.6} & 62.7 & 48.2 & 85.9 & 94.0 & 90.1 & 41.7          & 89.1 & 95.3 & \textbf{95.8} & \textbf{61.8} \\
				grid        & \textbf{97.4} & \textbf{98.9} & \textbf{89.8} & \textbf{32.2} & 96.0 & 98.0 & 71.8 & 29.5 & 81.3 & 89.9 & 77.3 & 8.7  & 86.5 & 93.5 & 87.4 & 24.3 \\
				hazelnut    & 95.9 & 97.5 & 88.5 & 70.0          & 98.0 & 98.0 & 92.7 & 61.6 & 99.1 & 99.2 & 97.5 & 69.6          & \textbf{99.2} & \textbf{99.4} & \textbf{99.1} & \textbf{77.5} \\
				leather     & 99.1 & 99.7 & 91.9 & \textbf{65.9} & \textbf{100.0} & \textbf{100.0} & 82.4 & 60.1 & 96.5 & 98.3 & 94.6 & 47.1 & 86.2 & 94.1 & \textbf{97.2} & 45.1 \\
				metal\_nut  & \textbf{99.8} & \textbf{99.9} & \textbf{98.7} & \textbf{96.4} & 99.5 & \textbf{99.9} & 98.3 & 89.1 & 73.8 & 90.0 & 83.2 & 38.5 & 85.6 & 95.0 & \textbf{92.2} & 69.1 \\
				pill        & 93.2 & 98.6 & 94.9 & \textbf{79.3} & \textbf{97.5} & \textbf{99.4} & \textbf{96.4} & 78.9 & 89.7 & \textbf{97.3} & 93.3 & 65.1 & 87.7 & 96.4 & 95.4 & 65.1 \\
				screw       & 12.5 & 53.7 & 56.8 & 1.7           & 37.3 & 69.1 & 63.7 & \textbf{7.7}  & 73.3 & 85.8 & \textbf{95.6} & 4.7  & \textbf{100.0} & \textbf{100.0} & 46.6 & 0.2 \\
				tile        & \textbf{100.0} & \textbf{100.0} & 96.6 & \textbf{91.0} & \textbf{100.0} & \textbf{100.0} & 90.9 & 72.0 & 99.9 & \textbf{100.0} & 87.0 & 51.9 & \textbf{100.0} & \textbf{100.0} & \textbf{98.7} & 88.3 \\
				toothbrush  & \textbf{95.8} & \textbf{98.2} & 73.2 & \textbf{27.9} & 93.6 & 97.0 & 60.6 & 22.5 & 89.2 & 94.1 & 78.9 & 12.9          & 90.0 & 95.3 & \textbf{86.6} & 19.9 \\
				transistor  & \textbf{99.9} & \textbf{99.8} & \textbf{94.3} & \textbf{81.2} & 98.5 & 95.8 & 66.0 & 38.6 & 78.8 & 76.8 & 58.9 & 14.9          & 84.6 & 81.3 & 62.8 & 17.0 \\
				wood        & 98.6 & 99.4 & \textbf{94.6} & \textbf{75.7} & \textbf{99.5} & \textbf{99.8} & 86.9 & 68.1 & 99.0 & 99.6 & 80.7 & 42.2          & 96.9 & 98.6 & 81.7 & 44.0 \\
				zipper      & \textbf{100.0} & \textbf{100.0} & \textbf{89.5} & \textbf{68.5} & 99.6 & 99.9 & 58.2 & 39.8 & 80.8 & 91.7 & 86.8 & 18.5          & 74.9 & 90.7 & 81.8 & 22.5 \\
				\midrule
				\textbf{Mean} & 88.8 & 94.5 & \textbf{87.4} & \textbf{58.4} & \textbf{92.0} & \textbf{96.3} & 76.7 & 50.0 & 84.7 & 91.3 & 84.9 & 30.5          & 89.7 & 94.8 & 85.4 & 41.5 \\
				\bottomrule
		\end{tabular}}
	\end{table}
	
	To further evaluate the upper-bound capability of DeltaDeno, we compare our zero-shot framework against representative Few-Shot Anomaly Generation (FSAG) methods. It is crucial to note that FSAG methods inherently enjoy a massive advantage as they are allowed to observe and fine-tune on a small pool of real anomalous target samples and pixel-accurate masks, which introduces partial data leakage from the benchmark. In contrast, our method operates under a strict zero-shot regime without any adaptation. 
	
	As indicated in Table~\ref{tab:mvtec_vs_fsag}, despite FSAG methods benefiting from the direct visual guidance of real downstream anomalies during fine-tuning, DeltaDeno demonstrates highly competitive performance. Specifically, when empowered by the advanced FLUX.1-dev~\cite{flux2024}  backbone, DeltaDeno achieves an overall image-level performance of \textbf{89.7\% AUC$_i$} and \textbf{94.8\% AP$_i$}, closely rivaling the supervised few-shot approaches. 
	
	More impressively, our method not only outperforms these few-shot baselines on several specific categories, but also achieves a remarkably higher average pixel-level AUROC (AUC$_p$) than the recent few-shot method AnoGen (85.4\% vs. 76.7\%). This empirical evidence thoroughly verifies that by decoupling explicit defect scaling from model training, our step-wise denoising attribution mechanism can effectively exploit rich text-to-image foundation priors to establish robust, precise, and transferable anomaly representations without any annotation costs. 
	
	\subsection{Dependence on SAM}
	To investigate the dependence of our method on the Segment Anything Model (SAM), we conduct an ablation study by replacing the SAM-derived foreground prior with an all-ones mask. This SAM-free configuration yields an image-level AUC/AP of 83.3\%/90.7\% and a pixel-level AUC/AP of 82.1\%/29.2\% on the MVTec AD dataset. While this represents a slight drop compared to our default setting equipped with SAM (84.7\%/91.3\%/84.9\%/30.5\%), the overall zero-shot performance remains highly competitive. We observe that the marginal degradation primarily stems from certain defects being synthesized in background regions, which misaligns with the natural distribution of real test defects that typically manifest on foreground objects. Nevertheless, these results confirm that DeltaDeno remains fully functional without SAM. Furthermore, as demonstrated in Fig.~4(a) of the main paper, our framework can flexibly accept arbitrary manual masks to achieve precise spatial location control.
	
	\subsection{Runtime and Memory Efficiency}
	For industrial data bootstrapping, generation efficiency is a practical consideration. We evaluate the computational cost of the full DeltaDeno pipeline—including the two-branch denoising, prompt refinement, and SAM preprocessing—and compare it against recent ZSAG baselines. Evaluated on a single NVIDIA GPU, DeltaDeno requires approximately 10.25 seconds per image with a peak memory footprint of 10.36 GiB. In contrast, the diffusion-based AnomalyAny is significantly heavier, requiring 67.4 s/img and 31.7 GiB of peak memory. Meanwhile, AnoStyler, which relies on a non-diffusion-based lightweight style transfer mechanism, takes 14.36 s/img with 5.8 GiB peak memory. These comparisons demonstrate that DeltaDeno achieves an excellent balance, delivering superior generation realism and localization precision while maintaining highly manageable runtime and memory costs.
	
	\subsection{Quantitative Threshold Analysis}
	
	\begin{table}[h]
		\centering
		\caption{\textbf{Quantitative threshold sensitivity analysis on MVTec AD}. $T_1$ and $T_2$ denote the $t_{mid}$ threshold and the last-step mask threshold, respectively. The default configuration is marked in bold.}
		\label{tab:threshold_ablation}
		\resizebox{0.9\linewidth}{!}{
			\begin{tabular}{lcccc|lcccc}
				\toprule
				$T_1$ / $T_2$ & AUC$_i$ & AP$_i$ & AUC$_p$ & AP$_p$ & $T_1$ / $T_2$ & AUC$_i$ & AP$_i$ & AUC$_p$ & AP$_p$ \\
				\midrule
				0.2 / 0.35 & 84.2 & 90.3 & 82.2 & 30.0 & 0.6 / 0.60 & 88.6 & 93.2 & 69.1 & 20.2 \\
				0.4 / 0.35 & 86.3 & 92.6 & 84.1 & 35.3 & 0.6 / 0.45 & 82.8 & 89.2 & 77.0 & 26.4 \\
				\textbf{0.6 / 0.35} & \textbf{84.7} & \textbf{91.3} & \textbf{84.9} & \textbf{30.5} & \textbf{0.6 / 0.35} & \textbf{84.7} & \textbf{91.3} & \textbf{84.9} & \textbf{30.5} \\
				0.8 / 0.35 & 84.8 & 89.9 & 71.4 & 21.9 & 0.6 / 0.15 & 87.3 & 91.7 & 79.6 & 29.3 \\
				\bottomrule
		\end{tabular}}
	\end{table}
	We supplement the qualitative threshold analysis presented in the main paper with a comprehensive quantitative evaluation. Table~\ref{tab:threshold_ablation} details the downstream detection performance on MVTec AD when systematically varying the early-stage mask threshold ($t_{mid}$ threshold, denoted as $T_1$) and the last-step mask threshold (denoted as $T_2$). 
	
	$T_1$ controls the extent of the intermediate editable region. When $T_1$ is excessively large, the latent inpainting region becomes overly restricted, preventing the full morphological development of the defect. Conversely, $T_2$ governs the final mask extraction process and primarily influences the false-positive/false-negative (FP/FN) trade-off in pixel-level evaluation metrics. Across diverse object categories, empirical results indicate that fixing $T_2=0.35$ yields generally stable and robust performance. Meanwhile, the optimal $T_1$ value typically resides between 0.4 and 0.6. Based on these observations, we adopt $T_1=0.6$ and $T_2=0.35$ as our default configuration to strike a reliable balance across diverse industrial scenarios.
	
	\section{Discussion on Evaluation Protocols}
	\label{app:discussion}
	\subsection{Generation Quality Evaluation}
	Although widely used in anomaly generation research, current evaluation metrics are not fully aligned with the characteristics of zero-shot anomaly generation (ZSAG) methods. Most quantitative protocols including IC-LPIPS-based diversity estimation, implicitly rely on real anomalous samples as anchors or reference clusters. This design introduces several limitations in the zero-shot setting.
	
	First, ZSAG methods such as ours do not observe any real anomalies during generation. Their synthesized defects arise purely from the generative prior learned by the diffusion model, rather than from the empirical distribution of a specific dataset such as MVTec AD. As a consequence, using real anomalous images as ``cluster centers'' for LPIPS grouping assumes a semantic and structural correspondence that may not exist. A high or low IC-LPIPS value in such a setting may therefore reflect the mismatch between diffusion priors and dataset-specific defects, rather than the intrinsic diversity of the generated anomalies.
	
	Second, diversity measured only within clusters anchored on real anomalies penalizes any novel defect patterns that are plausible under the generative model but absent in the dataset. Such patterns are common in zero-shot generation and may even be beneficial for training detection models. However, IC-LPIPS anchored on real samples tends to assign them to distant or sparse clusters, artificially lowering the diversity score.
	
	Third, these metrics implicitly conflate ``distance to real defects’’ with ``generation quality.’’ In zero-shot regimes, the goal is not to reproduce dataset-specific anomalies, but to produce semantically coherent and structurally localized defects that are useful for downstream detectors. Metrics tied too closely to real anomaly manifolds fail to capture this more general objective.
	
	Overall, while existing metrics provide a convenient quantitative proxy, they do not fully capture the behavior or intended purpose of zero-shot anomaly generation. Developing evaluation protocols that directly assess generative realism, spatial coherence, and usefulness for downstream detection, without assuming access to real anomalies, remains an important direction for future work.
	
	\subsection{Downstream Detection Evaluation}
	\label{app:discussion-det}
	
	Most existing works assess the effectiveness of anomaly generation by training a downstream detector on the synthesized anomalies and then evaluating on standard benchmarks such as MVTec AD. In practice, the generated data are fed into diverse anomaly detection frameworks, ranging from U-Net-based segmentation models as in AnomalyDiffusion to the AnomalyGPT~\cite{anomalygpt} in AnomalyAny. The resulting image-level and pixel-level AUROC or AP scores are then used to compare different generation methods. 
	
	Regarding the evaluation protocol for the number of generated samples, prior FSAG works typically synthesize 500--1000 anomalous images per category from a large pool of normal images. In contrast, AnomalyAny adopts a more restrictive one-shot setting, where anomaly generation is conditioned on the same single normal sample and only 100 random anomalous samples are synthesized for downstream training. In this work, we follow the latter protocol. This setting intentionally limits both the number of required normal inputs and the amount of generated anomalies, aiming to reduce the computational and training cost of the downstream AD task while better reflecting practical deployment scenarios.
	
	While this protocol is intuitive, it has several limitations for evaluating zero-shot anomaly generation. First, it does not provide a fair comparison between zero-shot and few-shot anomaly generation. Few-shot methods are allowed to see a subset of real anomalous samples during training, either in the generator or in the detector, which effectively leaks label information from the target benchmark. Their downstream performance therefore reflects a mixture of direct supervision on real defects and the contribution of synthetic data, whereas zero-shot methods rely purely on pretrained priors without access to real anomalies.
	
	Second, the evaluation entangles the quality of the generator with the choice of detection architecture, optimization scheme, and hyperparameters. Different works adopt different backbones, loss functions, and training recipes. As a result, performance gains or drops may be driven by how well a particular detector fits the generated distribution rather than by the intrinsic usefulness of the synthetic anomalies. High-capacity or strongly regularized detectors can partially compensate for low-quality synthetic data, and fragile detectors can mask improvements brought by better generators.
	
	Third, training a dedicated detector on generated data for a single benchmark tends to overestimate dataset-specific adaptation and underestimate generalization. The metric primarily measures how well the generation pipeline supports a particular detection architecture on MVTec AD, rather than how well it captures generic, transferable defect priors. This is especially problematic for zero-shot methods, whose goal is to leverage broad pretrained knowledge rather than to approximate the exact anomaly manifold of one dataset.
	
	Overall, downstream detection remains an important proxy for practical utility, but current protocols are still coarse and confounded. More principled evaluation would control the detection backbone and training settings across methods, measure relative improvements over a detector trained without synthetic anomalies, and include cross-dataset transfer to reduce dependence on a single benchmark.
	
	\subsection{Mask--Image Pair Quality Evaluation}
	\label{app:discussion-mask}
	
	Most anomaly generation studies evaluate only the realism of synthesized images, even though anomaly generation is inherently a paired task that requires both an anomalous image and a corresponding defect-region mask. Assessing image quality alone ignores the quality of the generated masks, which is crucial because the mask specifies where the model believes the synthesized defect is located.
	
	The primary aspect of mask quality is its spatial alignment with the generated anomaly. Existing protocols typically capture this only indirectly through downstream anomaly detection performance, which mixes mask accuracy with detector design, training configuration, and synthetic-to-real domain gaps. In addition, factors such as mask resolution, boundary sharpness, and the consistency between mask shape and defect semantics also influence practical usefulness. A more complete evaluation of anomaly generation should therefore consider both image fidelity and mask accuracy, rather than relying solely on image-based metrics or downstream detection scores.

	\section{Extended Qualitative Visualizations}
	\label{app:vis}
	
\begin{figure*}[t]
		\centering
		\includegraphics[width=\linewidth]{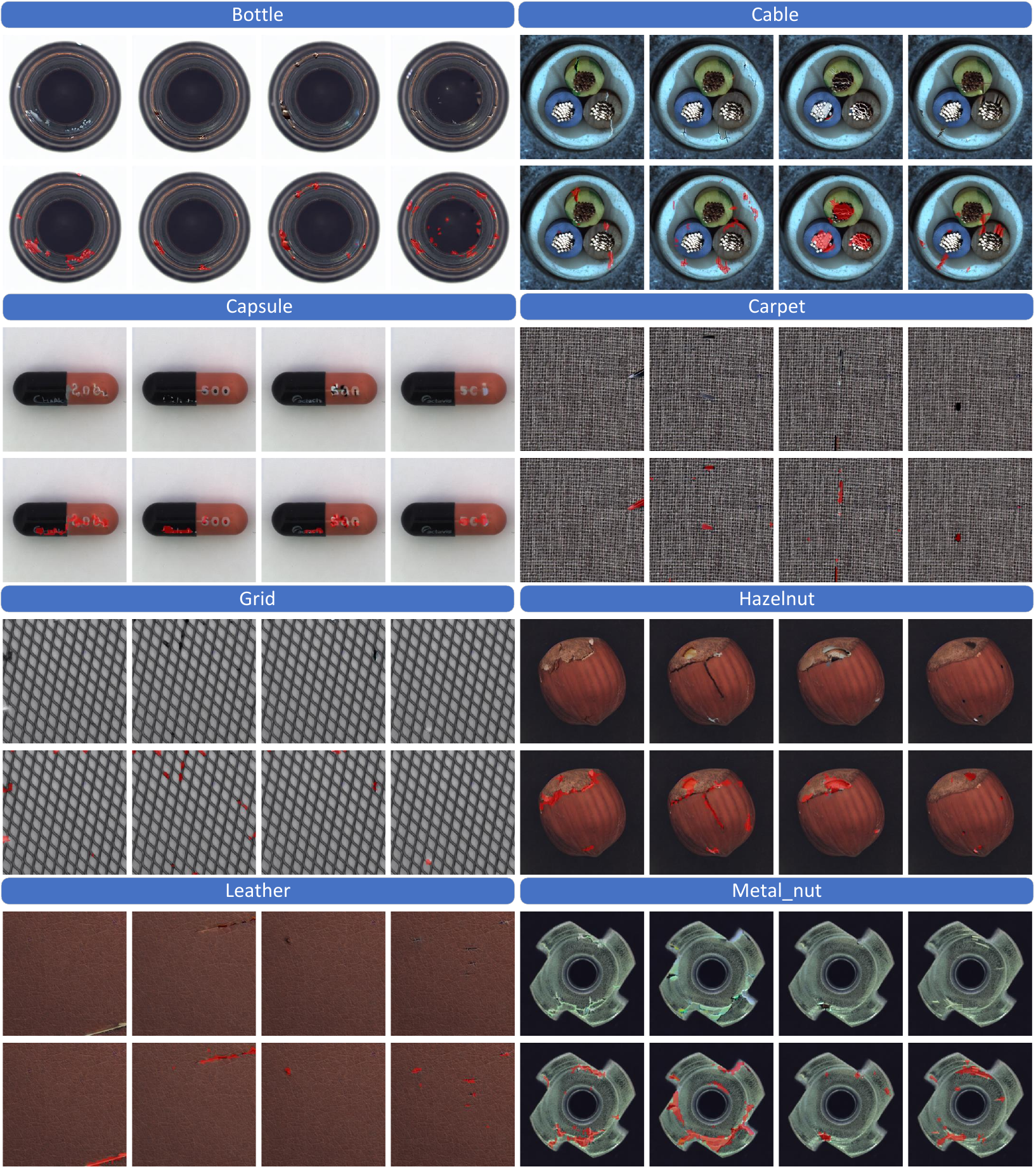}
		\caption{\textbf{Qualitative visualization of generated anomalies.} We present synthesized defect images and their paired masks across various object categories.}
		
		\label{fig:vis1}
	\end{figure*}

\begin{figure*}[t]
		\centering
		\includegraphics[width=\linewidth]{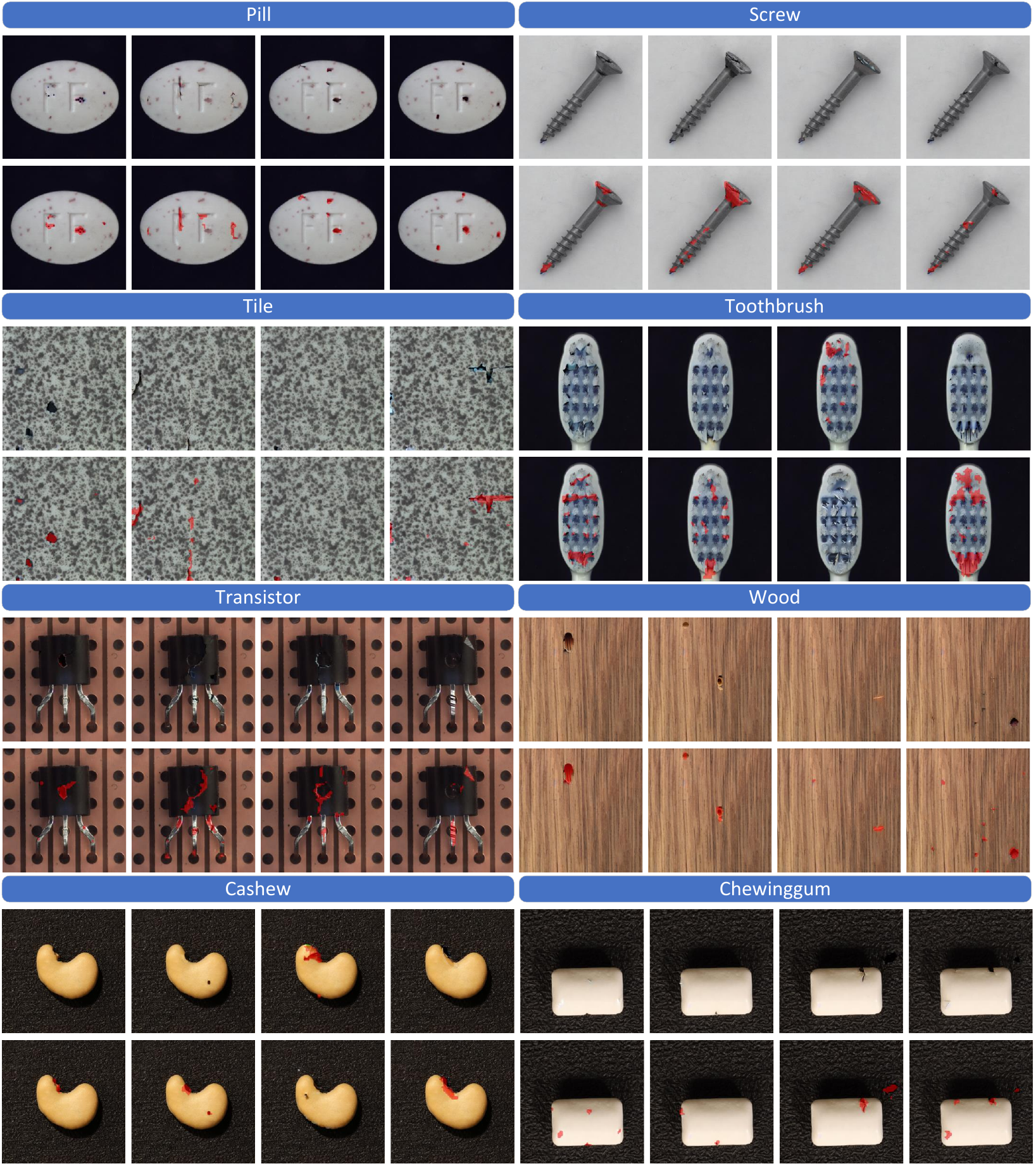}
		\caption{\textbf{Qualitative visualization of generated anomalies.} We present synthesized defect images and their paired masks across various object categories.}
		
		\label{fig:vis2}
	\end{figure*}

	In this section, we provide extensive qualitative examples to further demonstrate the capabilities of DeltaDeno. Fig.~\ref{fig:vis1} and Fig.~\ref{fig:vis2} showcase a wide variety of synthesized anomalous images paired with their corresponding generated masks across multiple industrial categories. 

\end{document}